%% file: AnonymousSubmission2027.tex
\documentclass[letterpaper]{article} % DO NOT CHANGE THIS
\usepackage[preprint]{aaai2027}  % DO NOT CHANGE THIS
\usepackage[hyphens]{url}  % DO NOT CHANGE THIS
\usepackage{graphicx} % DO NOT CHANGE THIS
\usepackage{natbib}  % DO NOT CHANGE THIS AND DO NOT ADD ANY OPTIONS TO IT
\usepackage{caption} % DO NOT CHANGE THIS AND DO NOT ADD ANY OPTIONS TO IT
\usepackage{amsmath}
\usepackage{amssymb}
\usepackage{comment}
 
\usepackage{algorithm}
\usepackage{algorithmic}

\usepackage{newfloat}
\usepackage{listings}
\DeclareCaptionStyle{ruled}{labelfont=normalfont,labelsep=colon,strut=off} % DO NOT CHANGE THIS
\floatstyle{ruled}
\newfloat{listing}{tb}{lst}{}
\floatname{listing}{Listing}

\usepackage{booktabs}

\title{Mode-Dependent Rectification for Stable PPO Training}

\author{
    Mohamad Mohamad\textsuperscript{\rm 1}\corresponding,
    Francesco Ponzio\textsuperscript{\rm 2},
    Xavier Descombes\textsuperscript{\rm 1}
}
\affiliations{
    \textsuperscript{\rm 1}I3S, Inria, Université Côte d'Azur, Nice, France\\
    \textsuperscript{\rm 2}Department of Control and Computer Engineering, Politecnico di Torino, Turin, Italy\\
    *mohamad.mohamad@inria.fr

}

\begin{document}

\maketitle

\begin{abstract}
Mode-dependent architectural components (layers that behave differently during training and evaluation, such as Batch Normalization or dropout) are commonly used in visual reinforcement learning but can destabilize algorithms such as Proximal Policy Optimization (PPO). Discrepancies between training and evaluation behavior induced by Batch Normalization can lead to policy mismatch, distributional drift, and reward collapse. We propose Mode-Dependent Rectification (MDR), a lightweight two-phase training procedure that stabilizes PPO under mode-dependent layers without architectural changes. Experiments across Procgen, Atari, and real-world patch-localization tasks demonstrate that MDR improves stability and performance of Batch Normalization, and extends to other mode-dependent layers.
\end{abstract}


\section{Introduction}
\label{sec:intro}
Much of deep learning’s progress has been driven by architectural innovations that enable large-scale optimization. Residual connections \cite{resnet}, transformer blocks \cite{attention}, and normalization layers \cite{batchnorm,layernorm,groupnorm} have each played a central role in shaping modern neural networks. Among these, Batch Normalization (BatchNorm) \cite{batchnorm} remains particularly influential, having improved optimization stability and training efficiency across a wide range of models. Despite the introduction of alternatives such as LayerNorm and GroupNorm \cite{layernorm,groupnorm}, BatchNorm remains widely used in convolutional architectures.

In reinforcement learning (RL), normalization behaves differently. Unlike supervised learning, where data are drawn from a stationary distribution, RL data are generated by an evolving policy, inducing non-stationary state distributions. As a result, layers such as BatchNorm, whose behavior depends on data-driven statistics, are sensitive to discrepancies between training and evaluation distributions.

Prior work has reported mixed findings on the use of BatchNorm in RL. In off-policy settings, large replay buffers aggregate data from multiple past policies, introducing inter-policy distribution shifts even within a single batch \cite{Crossnorm,CrossQ}. For PPO \cite{schulman2017ppo}, one might expect BatchNorm to behave more consistently, since training data are generated by the current policy. Yet empirical evidence remains mixed, with reports of both performance gains and degradation  \cite{CoinRun,Rogue}.

In our experiments, incorporating BatchNorm into PPO resulted in catastrophic reward collapse. Stable training could be recovered by fixing BatchNorm layers in evaluation mode using pretrained (ImageNet) statistics. 

These observations motivate the central question of this work:
\emph{can the instability induced by mode-dependent layers such as BatchNorm be systematically corrected, and is doing so beneficial for policy performance?}

We decompose this into four questions: (i) \emph{why} does BatchNorm destabilize PPO; (ii) can this instability be corrected; (iii) does the correction extend to other mode-dependent layers such as dropout; and (iv) is the correction worthwhile, that is, is a corrected layer competitive with schemes that sidestep the failure entirely?

We first analyze the mechanism underlying BatchNorm-induced reward collapse in PPO, arguing that instability arises from a mismatch between training- and evaluation-mode policies amplified by distributional shift. We generalize this analysis to mode-dependent layers and propose Mode-Dependent Rectification (MDR), a two-phase training procedure that interleaves standard optimization with a deterministic rectification phase. We evaluate MDR with PPO on Procgen, Atari, and two real-world patch-localization benchmarks using two backbone architectures. Our results show that MDR consistently stabilizes mode-dependent layers wherever instability arises: it restores stable training and achieves the strongest performance where BatchNorm collapses while closely matching BatchNorm elsewhere. A corrected BatchNorm remains competitive with normalization schemes that avoid the train--evaluation mismatch, and the same rectification procedure also stabilizes dropout. Our main contributions are as follows:

\begin{itemize}
    \item We analyze how BatchNorm induces reward collapse in PPO through policy mismatch and distributional shift, and extend this analysis to the broader class of mode-dependent layers.
    
    \item We propose Mode-Dependent Rectification (MDR), a simple two-phase procedure that stabilizes optimization without architectural changes.
    
    \item We evaluate MDR across three benchmarks and two architectures, showing that a corrected mode-dependent layer is competitive with alternatives that avoid the failure.
\end{itemize}

\section{Background}
\label{sec:background}
\subsection{Normalisation}

Normalization layers are now standard components in deep neural networks, largely due to their ability to stabilize optimization and reduce sensitivity to parameter initialization. BatchNorm \cite{batchnorm} introduced the use of minibatch statistics together with running estimates for inference, substantially accelerating training but making performance dependent on i.i.d. batches and sufficiently large batch sizes. Batch Renormalization \cite{batchrenorm} alleviates this limitation by interpolating between batch statistics and running averages, improving robustness when batch distributions drift or batch sizes are small.

A range of alternatives were subsequently proposed to remove this dependence on batch composition. Layer Normalization \cite{layernorm} computes per-sample statistics across all features, making it agnostic to batch size. Instance Normalization \cite{instance-norm} normalizes each channel independently for every sample. Group Normalization \cite{groupnorm} partitions channels into groups and computes statistics within each group, remaining stable even for very small minibatches.

More recently, CrossNorm and SelfNorm \cite{cross-selfnorm} were introduced to improve generalization under distribution shift by manipulating channel-wise feature statistics. CrossNorm enlarges the effective training distribution by exchanging channel-wise means and variances between feature maps. In contrast, SelfNorm learns to recalibrate these statistics through attention mechanisms. CrossNorm is applied only during training.

\begin{figure}[t]
  \centering
\includegraphics[width=0.9\linewidth,keepaspectratio]{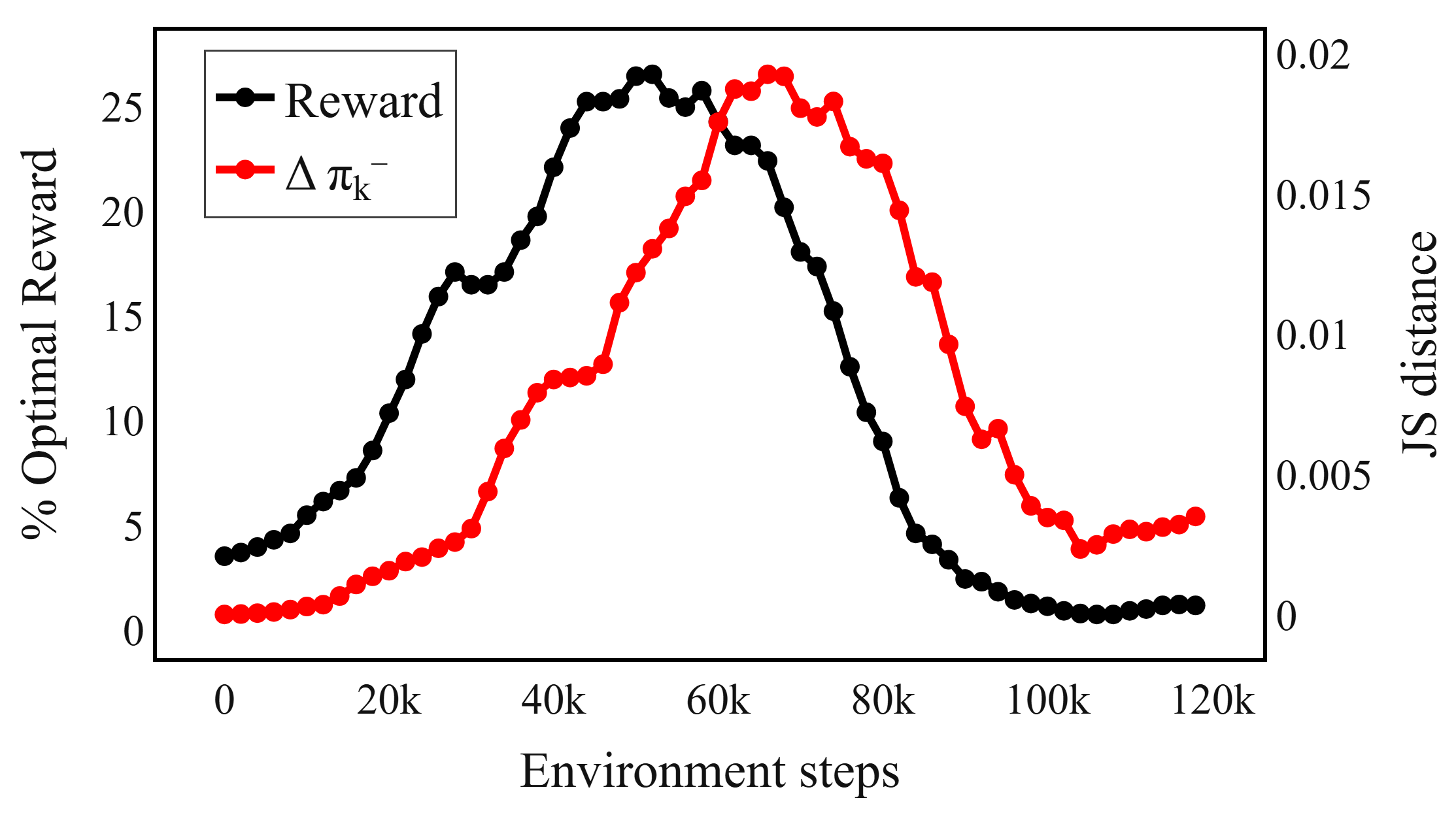}
    \caption{Reward collapse. Training curves when BatchNorm operates in training mode during optimization and evaluation mode during data collection. Rewards initially improve rapidly; as training progresses, the mismatch $\Delta \pi_k^-$ increases. Beyond a critical point, this growing mismatch coincides with a sudden collapse in performance.}

    \label{fig:collapse}
\end{figure}

\subsection{Normalisation and Dropout in RL}

The incorporation of normalization and mode-dependent layers in RL has received increasing attention in recent years. DroQ ~\cite{dropQ} introduced the use of dropout layers together with layer normalization to regularize and “shrink’’ the Q-function ensemble used in REDQ \cite{REDQ} —the state-of-the-art at the time for sample efficiency through the usage of a large ensemble of Q functions. While DroQ maintains sample efficiency close to REDQ, it achieves substantially better computational efficiency, comparable to that of SAC \cite{SAC}.

Another line of work studies the role of normalization in off-policy algorithms. \cite{Crossnorm} showed that naïvely applying BatchNorm is ineffective due to the action-distribution shift between the critic and the target networks (one updated under past policies, the other under the current policy). They proposed normalizing features on a mixture of on-policy and off-policy samples, combined with removing the target network, yielding significant improvements in TD3 ~\cite{td3} and DDPG ~\cite{DDPG}. Building on this, they later introduced CrossQ  ~\cite{CrossQ}, an off-policy method that uses wider critic networks and carefully applied BatchNorm and Batch Renormalization, and reported performance surpassing both DroQ and REDQ.

CrossQ+WN \cite{CrossQ_WN} then examined why CrossQ fails to scale to higher UTD ratios, attributing it to the scale-invariance of BatchNorm layers, which allows parameter norms to grow and the effective learning rate to drift. They propose weight normalization~\cite{NELR} as a simple remedy that restores stable scaling to higher UTDs. Building further on this, XQC \cite{XQC} analyze the critic's optimization landscape through the eigenspectrum and condition number of its Hessian, and show that combining batch normalization, weight normalization, and a distributional cross-entropy critic loss better conditions the loss and improves performance and robustness. In their analysis and ablations, BatchNorm proves particularly important, driving gains in both conditioning and performance over dense and LayerNorm baselines.

Beyond sample efficiency, \cite{NEG} investigated the use of CrossNorm and SelfNorm for improving generalization in RL, showing dramatic gains—for example, an increase from 14\% to 97\% success on CARLA. In contrast, most work in this area focuses on off-policy RL; results in on-policy settings remain mixed.  \cite{CoinRun} demonstrated that BatchNorm and dropout can improve generalization on the CoinRun benchmark \cite{CoinRun,Proce}. However, \cite{Rogue} found the opposite on their Rogue benchmark: BatchNorm underperformed relative to architectures that did not employ it, whereas LayerNorm performed better. Additional work \cite{REPPO,NELR,Plasticity} has further explored the role of LayerNorm in mitigating plasticity loss, reviving dormant neurons, and improving network stability.

To the best of our knowledge, no prior work has explained the discrepancy between these conflicting findings for BatchNorm in on-policy RL. Our work provides the first explanation of the underlying mechanism through which BatchNorm harms PPO in on-policy settings.

\section{Method} 
\label{sec:method}

\subsection{Preliminary}

\noindent\textbf{Batch Normalization.}
BatchNorm \cite{batchnorm} mitigates covariate shift in deep neural networks by normalizing activations within each mini-batch. Formally, given a mini-batch $B = \{x_1, x_2, \dots, x_m\}$ of size $m$, the batch mean and variance are computed as:
\begin{equation}
\mu_B = \frac{1}{m} \sum_{i=1}^{m} x_i,
\quad
\sigma_B^2 = \frac{1}{m} \sum_{i=1}^{m} (x_i - \mu_B)^2
\label{eq:batch stats}
\end{equation}
The normalized activations are
\begin{equation}
\hat{x}_i = \frac{x_i - \mu_B}{\sqrt{\sigma_B^2 + \eta}}, \quad
y_i = \gamma \hat{x}_i + \beta, \quad i = 1, \dots, m
\label{eq:normalised activation}
\end{equation}
where $\eta$ is a small constant for numerical stability, $\gamma$ and $\beta$ are learnable parameters. At inference, when batch statistics are unavailable, BatchNorm uses global estimates of the mean and variance, $\mu_{\text{global}} = \mathbb{E}[D]$ and $\sigma_{\text{global}}^2 = \mathbb{V}[D]$ approximated during training via running averages with momentum $\mathcal{M}$:
\begin{equation}
\begin{aligned}
\mu_{\text{r}} &= (1 - \mathcal{M}) \, \mu_{\text{r}} + \mathcal{M} \, \mu_B, \\
\sigma_{\text{r}}^2 &= (1 - \mathcal{M}) \, \sigma_{\text{r}}^2 + \mathcal{M} \, \sigma_B^2
\end{aligned}
\label{eq:running_stats}
\end{equation}
\noindent\textbf{Proximal Policy Optimization.}
Proximal Policy Optimization (PPO) \cite{schulman2017ppo} optimizes a clipped surrogate objective:
\begin{equation}\label{eq:clip}
\begin{split}
L^{\text{CLIP}}(\theta)
&= \mathbb{E}_t\!\left[
      \min\!\Big(
        r_t(\theta)\hat{A}_t,\,
        \operatorname{clip}(r_t(\theta),1-\epsilon,1+\epsilon)\hat{A}_t
      \Big)
    \right] \\[6pt]
&\quad \quad \quad \text{with}\quad
r_t(\theta)
= \frac{\pi_\theta(a_t\,|\,s_t)}{\pi_{\theta_{\text{old}}}(a_t\,|\,s_t)}
\end{split}
\end{equation}
Here $\hat{A}_t$ is an advantage estimate and $\pi_\theta$ is a stochastic policy. The clipping term prevents large policy updates, restricting changes to a trust region defined by $\epsilon$. The full PPO objective also includes a value-function loss and an entropy bonus:
\begin{equation}\label{eq:ppo}
\begin{split}
L^{\text{PPO}}(\theta) &=
\mathbb{E}_t \Big[
L^{\text{CLIP}}_t(\theta)
- c_1 \, L^{\text{VF}}_t
+ c_2 \, S[\pi_\theta](s_t)
\Big]  \\[6pt]
&\quad \quad L^{\text{VF}}_t=(V_w(s_t) - V_t^{\text{target}})^2
\end{split}
\end{equation}
where $V_w$ is the critic, $S$ denotes the entropy bonus, and $c_1, c_2$ are weighting coefficients that balance the value and entropy terms.

\subsection{BatchNorm Induced Instability}
\label{subsec:Observation}
\noindent\textbf{Motivation.}
During preliminary experiments with ResNet18, we found that using BatchNorm layers in the conventional supervised learning manner (by setting them to training mode during policy updates and evaluation mode during interaction) led to severe training instability. In some environments, the agent's reward collapsed rapidly. Stability was restored when BatchNorm was fixed to evaluation mode, using running statistics computed from ImageNet. This unexpected behavior motivated a systematic investigation into the interaction between BatchNorm and policy optimization.
 
\noindent\textbf{Formulation of the Phenomenon.}
We define the following notation. Let $k$ index training steps, each consisting of a data collection phase followed by multiple parameter update iterations. Let $\pi_{\theta_k}$ denote the policy at step $k$, $\mu_k$ the corresponding state distribution, and $D_k \sim \mu_k$ the dataset collected at that step. Superscripts $-$ and $+$ indicate quantities before and after training within a step, respectively.
 
When BatchNorm operates in training mode, it uses mini-batch statistics $(\mu_B, \sigma_B)$, inducing the policy $\pi^{\mu_B, \sigma_B}_{\theta_k}$. In evaluation mode, BatchNorm relies on running statistics $(\mu_r, \sigma_r)$, yielding $\pi^{\mu_r, \sigma_r}_{\theta_k}$. During each step $k$, the dataset $D_k$ is collected under the evaluation-mode policy $\pi^{\mu_r, \sigma_r}_{\theta_k}$.
We can then define the mismatch between the two BatchNorm mode policies as:
\begin{equation}
\Delta \pi_{k}^- = \mathbb{E}_{s \sim \mu_k}\!\left[ d\big(\pi^{\mu_B, \sigma_B}_{\theta_k}(\cdot \mid s), \pi^{\mu^-_r, \sigma^-_r}_{\theta_k}(\cdot \mid s)\big)\right]
\label{eq:shift_before}
\end{equation}
where $d(\cdot, \cdot)$ represents a probability distribution distance. After training, both the policy parameters and running statistics are updated, leading to
\begin{equation}
\Delta \pi_{k}^+ = \mathbb{E}_{s \sim \mu_k}\!\left[ d\big(\pi^{\mu_B, \sigma_B}_{\theta_{k+1}}(\cdot \mid s), \pi^{\mu^+_r, \sigma^+_r}_{\theta_{k+1}}(\cdot \mid s)\big)\right]
\label{eq:shift_after}
\end{equation}
While the above expressions describe a single step, training unfolds over multiple steps, during which mismatches compound recursively. In practice, $\Delta \pi_{k}^+$ propagates into $\Delta \pi_{k+1}^-$.
 
As training progresses, the policy $\pi_{\theta_k}$ evolves, inducing a corresponding shift in the state distribution. While early datasets satisfy $\mu_0 \approx \mu_1$, later distributions diverge, so that $\mu_k \not\approx \mu_{k+1}$. As a result, the running statistics (computed incrementally up to step $k$) no longer match the true batch statistics of the current step. Formally:
\begin{equation}
\mathbb{E}_{X \sim \mu_{k+1}}[X] \not\approx \frac{1}{k+1} \sum_{i=0}^{k} \mathbb{E}_{X \sim \mu_i}[X]\,,
\label{eq:stats_misalignment}
\end{equation}
This amplifies the discrepancy $\Delta \pi_{k}^-$ at the beginning of each 
training cycle. Under sufficiently severe shift, the model may exhibit training collapse, although the precise behavior is problem-dependent.
 
When policy updates are aggressive, they can temporarily degrade performance. While PPO's clipped objective can partially correct such updates via negative advantages, collapse is often accompanied by a substantial distribution shift between pre- and post-collapse rollouts (i.e., $\mu_{\text{pre-collapse}}$ and $\mu_{\text{post-collapse}}$). This shift amplifies instability, as well as the mismatch $\Delta \pi_k^-$, and hinders recovery.

We illustrate this behavior in Figure~\ref{fig:collapse} by estimating $\Delta \pi_{k}^-$ as the average distance $d(\cdot, \cdot)$ computed over all mini-batches within each dataset $D_k$. For the distance metric, we employ the Jensen--Shannon (JS) divergence, which provides a symmetric and bounded alternative to the Kullback--Leibler (KL) divergence. As shown in the figure, the value of $\Delta \pi_{k}^-$ increases concurrently with the agent's reward, indicating that the statistical mismatch between BatchNorm modes grows as the policy improves. Eventually, this accumulation of mismatch leads to policy collapse, after which the divergence remains high due to the pronounced distributional shift between pre-collapse and post-collapse rollouts.
 
\begin{figure*}[t]
  \centering
  \includegraphics[width=0.99\linewidth,keepaspectratio]{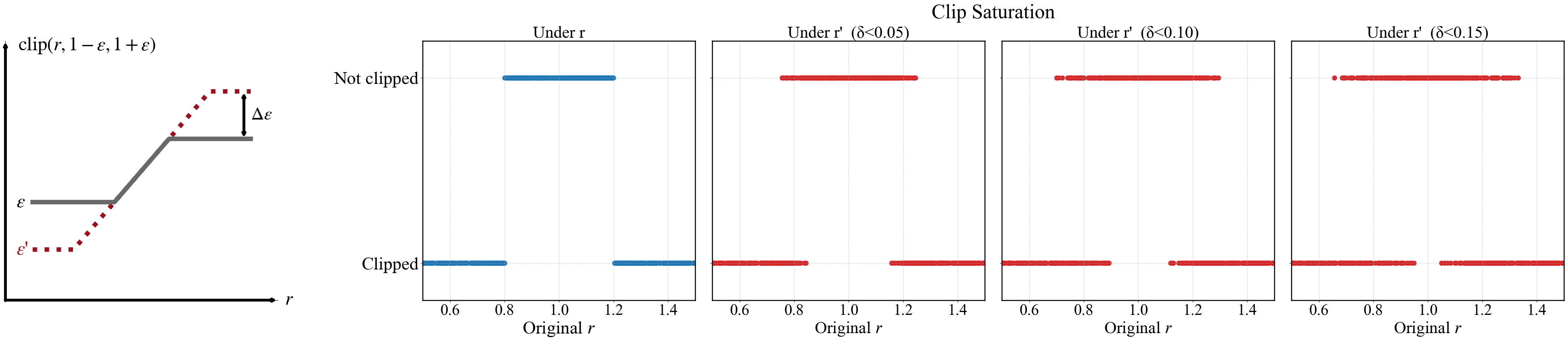}
  \caption{Effect of $\delta r$ and $\Delta\epsilon$ on PPO clipping. \textbf{Left}: Illustration of how a perturbation $\Delta \epsilon$ enlarges the effective clipping range of the PPO objective. \textbf{Right}: Clipping saturation as a function of the original ratio $r$. Blue points correspond to clipping under the unperturbed ratio $r$ ($\delta r = 0$), while red points show clipping under the perturbed ratio $r'$ with bounded noise $\delta r \in \{0.05, 0.10, 0.15\}$. Increasing $\delta r$ progressively expands the effective clipping boundaries.}
  \label{fig:clipping_and_saturation}
\end{figure*}
 
\subsection{General Formulation}
 
\noindent\textbf{Effect of Mode-Dependent Layers on the Policy Update.}
 
Let $N_\theta$ denote the policy network with parameters $\theta$, whose output defines the stochastic policy $\pi_\theta(\cdot \mid s)$. Suppose that certain layers within $N_\theta$ exhibit different behaviors between training and evaluation modes. This discrepancy induces a shift in the effective policy distribution at the level of individual states, which we define as:
\begin{equation}
\Delta \pi_\theta(s) = d\big(\pi^{\text{train}}_\theta(\cdot \mid s), \pi^{\text{eval}}_\theta(\cdot \mid s)\big)
\label{eq:general_shift}
\end{equation}
This state-wise formulation provides a state-level view of the step-level mismatch introduced in Eq.~\ref{eq:shift_before}. During the interaction phase, rollouts are collected and the action probabilities $\pi_\theta(a_t\,|\,s_t)$ are computed under the evaluation-mode policy $\pi^{\text{eval}}_\theta$, whereas during training, parameter updates are performed using the training-mode policy $\pi^{\text{train}}_\theta$. A direct consequence of $\Delta \pi_\theta$ is a shift between $\pi^{\text{train}}_\theta(a_t\,|\,s_t)$ and $\pi^{\text{eval}}_\theta(a_t\,|\,s_t)$ during optimization, which we denote by $\delta \pi_\theta(a_t\,|\,s_t)$.
 
Consequently, in the PPO surrogate objective, only the ratio term is affected as follows:
\begin{equation}
  \label{eq:r}
\begin{split}
r'_t(\theta)
&=
\frac{\pi^{\text{train}}_\theta(a_t\,|\,s_t)}
     {\pi^{\text{eval}}_{\theta_{\text{old}}}(a_t\,|\,s_t)} \\[6pt]
&=
\frac{\pi^{\text{eval}}_\theta(a_t\,|\,s_t) + \delta \pi_\theta(a_t\,|\,s_t)}
     {\pi^{\text{eval}}_{\theta_{\text{old}}}(a_t\,|\,s_t)}\\[6pt]
&=
r_t(\theta) + \delta r
\end{split}
\end{equation}
where $r'_t$ is the ratio actually used during optimization, $r_t$ is the ratio assuming mode-invariant behavior, and $\delta r$ captures the perturbation induced by mode-dependent layers.
 
We can interpret the consequences of this perturbation as follows:
 
\noindent\textbf{Trust-region shrinkage.}
When the perturbation $\delta r$ tightens the clipping range, it effectively shrinks the trust region (Figure~\ref{fig:clipping_and_saturation}, clipped region). In this regime, the PPO objective still enforces conservative updates, and learning remains stable.
 
\noindent\textbf{Trust-region violation and instability.}
In this case, the induced perturbation $\delta r$ results in policy updates that push $\pi_\theta$ outside the implicit trust region defined by $\epsilon$ in the clipped objective (Figure~\ref{fig:clipping_and_saturation}, unclipped region). We propose viewing this as a dynamic perturbation of the clipping boundary:
\begin{equation}
\epsilon' = \epsilon + \Delta \epsilon
\label{eq:eps_perturbation}
\end{equation}
where $\Delta \epsilon > 0$ captures the deviation induced by $\delta r$. Under this interpretation, the instability introduced by different mode-dependent layers manifests as a perturbation of the clipping parameter $\epsilon$, as if the policy were optimized under a stochastic distortion of the trust region.
Thus, $\Delta \epsilon$ captures the policy-level effect of mode-dependent layers without modeling each one explicitly. When it becomes sufficiently large, the trust-region guarantee of PPO is violated, often resulting in unstable learning dynamics or catastrophic policy collapse.

\begin{algorithm}[tb]
  \caption{PPO with MDR}
  \label{alg:ppo_alignment}
  \begin{algorithmic}
    \STATE {\bfseries Input:} policy $\pi_\theta$, value function $V_w$, mini-batch size $m$, phase coefficients $\alpha_1, \alpha_2$
    \STATE Initialize policy parameters $\theta$ and value parameters $w$
    
    \FOR{each training step $k = 1, 2, \ldots, K$}
    
      \STATE {\bfseries Interaction Phase:}
      \STATE Collect rollouts $\mathcal{D}_k$
      \STATE Compute advantages $\hat{A}_k$ and value targets $\hat{V}_k$
      
      \STATE {\bfseries Training Phase:}
      \STATE Set network to \texttt{train} mode
      \FOR{$i = 1$ {\bfseries to} $\alpha_1 \times \frac{|\mathcal{D}_k|}{m}$}
        \STATE Sample mini-batch $B \sim \mathcal{D}_k$
        \STATE Update $\theta, w$ using PPO objective $L^{\text{PPO}}$
      \ENDFOR
      
      \STATE {\bfseries Rectification Phase:}
      \STATE Set network to \texttt{eval} mode
      \FOR{$j = 1$ {\bfseries to} $\alpha_2 \times \frac{|\mathcal{D}_k|}{m}$}
        \STATE Sample mini-batch $B \sim \mathcal{D}_k$
        \STATE Update $\theta, w$ using PPO objective $L^{\text{PPO}}$
      \ENDFOR
      
    \ENDFOR
  \end{algorithmic}
\end{algorithm}

\subsection{Rectification}

The preceding analysis identifies the source of instability as a perturbation $\Delta\epsilon$ of PPO's trust region. Importantly, MDR does not attempt to estimate or compensate for $\Delta\epsilon$ directly. Since train--evaluation discrepancies differ across layers, doing so would require layer-specific assumptions. Instead, MDR corrects their shared effect at the policy level. Because the policy that collects the next rollout is the deterministic (evaluation-mode) policy, MDR periodically re-optimizes under deterministic layer behavior ($\Delta\epsilon=0$), re-anchoring the collection policy to PPO's original trust region before environment interaction resumes.
 
MDR modifies the training procedure by splitting each optimization cycle into two phases:
\begin{enumerate}
    \item \textbf{Standard update phase.} This corresponds to conventional training, performed for $\alpha_1 \times \frac{\mathrm{card}(D_k)}{m}$ iterations, where $\mathrm{card}(D_k)$ denotes the dataset size and $m$ the mini-batch size.
    \item \textbf{Rectification phase.} An auxiliary training phase in which all layers are switched to their deterministic (evaluation) mode and optimized for $\alpha_2 \times \frac{\mathrm{card}(D_k)}{m}$ iterations, where $\alpha_1$ and $\alpha_2$ are tunable hyperparameters satisfying $\alpha_1 + \alpha_2 = n$, the fixed per-rollout epoch budget.
\end{enumerate}

Rectification is performed immediately before the next round of experience collection, after deviations from PPO's original trust region have accumulated during the standard update phase. This ensures that interaction always resumes from a policy re-anchored to the intended trust region.

\section{Experiments}
\label{sec:result}
 
\begin{figure*}[t!]
  \centering
\includegraphics[width=0.99\linewidth,keepaspectratio]{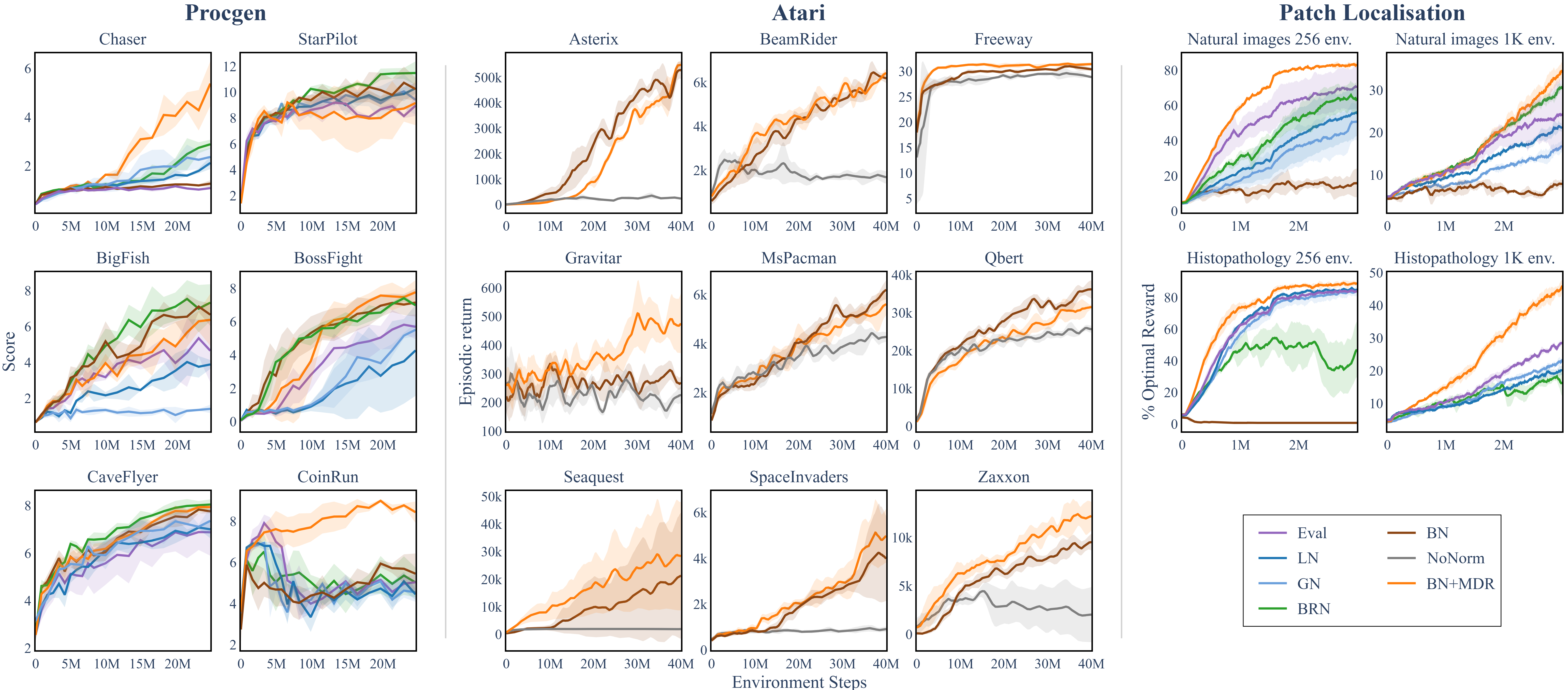}

\caption{Performance comparison across Procgen (left), Atari (center), and
    patch-localization (right) benchmarks. Procgen: average episode return
    (Score) on six games. Atari: episodic return on nine games with sticky actions. Patch localization: normalized reward, expressed as a percentage of the optimal policy reward, on the natural-image and  histopathology tasks with 256 and 1024 environments. Shaded regions denote one standard deviation across three seeds.}

    \label{fig:graph}
\end{figure*}

\subsection{Experimental Setting}

We evaluate MDR on two widely used mode-dependent layers, BatchNorm and dropout, across three benchmarks spanning both game-based and visual tasks.

Our evaluation includes six games from the Procgen benchmark \cite{Proce}: CoinRun, Chaser, StarPilot, BigFish, BossFight, and CaveFlyer; nine Atari games from the Arcade Learning Environment (ALE) \cite{ale,ale2} using sticky actions: Asterix, BeamRider, Freeway, Gravitar, MsPacman, Qbert, Seaquest, SpaceInvaders, and Zaxxon; and a patch-localization benchmark \cite{bioimaging25} consisting of a histopathology task together with a natural-image variant constructed from high-resolution OpenImages images \cite{openimage}. Procgen and Atari provide diverse control and visual domains, while Procgen further introduces substantial visual variability through procedural generation. We evaluate a representative subset of environments from both benchmarks, as evaluating every game is computationally prohibitive. The patch-localization benchmark provides a complementary setting, exhibiting severe BatchNorm instability and pronounced overfitting, making it well suited for evaluating both BatchNorm and dropout.

For Procgen and patch localization, we employ a shallow ResNet-18 backbone obtained by removing the final residual block and initialize all models using ImageNet-pretrained weights, which consistently improved performance across all normalization variants. Atari experiments instead use an enlarged IMPALA-CNN architecture based on the CleanRL implementation \cite{cleanrl}, increasing both network depth and width. Employing different architectures across benchmarks provides a broader evaluation of MDR.

For BatchNorm, the Procgen and patch-localization benchmarks compare six variants: (i) BN, where BatchNorm layers operate in standard training mode during training; (ii) Eval, where BatchNorm layers remain in evaluation mode throughout training; (iii) BN+MDR, our proposed two-phase training procedure; and three alternative normalization schemes, Layer Normalization (LN), Group Normalization (GN), and Batch Renormalization (BRN), obtained by directly replacing the BatchNorm layers in the network backbone.

For the Atari benchmark, we instead restrict the comparison to BN, BN+MDR, and the standard IMPALA architecture without normalization (NoNorm). As Atari experiments are substantially more computationally demanding, our objective is to evaluate possible benefits of MDR relative to the standard BatchNorm implementation and the baseline architecture, rather than perform an exhaustive comparison of normalization techniques.

For dropout, we compare three architectures: (i) NoNorm, where all normalization layers are removed; (ii) Dropout, where BatchNorm is replaced by dropout and trained conventionally; and (iii) Dropout+MDR, where the same architecture is trained using the proposed rectification phase.

Unless otherwise stated, all MDR experiments evaluate two allocations of the epoch budget $n$ between standard optimization and rectification: $(\alpha_1, \alpha_2)=(n{-}1,1)$ and $(\alpha_1, \alpha_2)=(1,n{-}1)$. These settings span light and extensive rectification within a fixed epoch budget. We report the better of the two allocations, with the full comparison in the appendix.

Procgen agents are trained for 25M steps on 500 easy levels. Atari agents are trained following the CleanRL implementation for 40M steps with sticky actions. BatchNorm experiments on patch localization are conducted for 3M steps using both the histopathology and natural-image variants with 256 and 1024 training environments. dropout experiments are performed on the histopathology task using 2048 training environments for 2.5M environment steps. Unless otherwise stated, all experiments are repeated over three random seeds.

In total, our experimental evaluation spans more than 350 GPU-days, primarily using NVIDIA A40 GPUs. Additional implementation details, hyperparameters, architectural choices, and benchmark details are provided in the appendix.

\subsection{BatchNorm with MDR}
\label{Subsec:BN-align}

We report the BatchNorm results across all three benchmarks in Figure~\ref{fig:graph}.

The right panel reports performance on the patch-localization tasks\footnote{A small performance increase mid-training is due to a learning rate scheduler used in the patch-localization experiments.}, expressed as a percentage of the optimal policy reward. BN+MDR consistently achieves the highest performance across both the natural-image and histopathology settings. BN fails on the natural-image task and is omitted for the 1024-environment histopathology setting due to complete reward collapse, already evident in the 256-environment experiment. Among the alternative normalization schemes, BRN improves over BN on natural images but performs poorly on histopathology, suggesting that while renormalization partially alleviates instability, rectification provides a more robust solution across domains.

On Atari (center) and Procgen (left), BN exhibits considerably less instability. It struggles mainly on three games: Chaser, CoinRun, and Gravitar, where BN+MDR consistently achieves the highest returns. Elsewhere, BN+MDR closely matches BN. Notable exceptions include StarPilot, where it underperforms, and Zaxxon and Freeway, where it respectively achieves higher returns and faster convergence. Across both benchmarks, BatchNorm-based models consistently outperform the corresponding architectures without BatchNorm, as well as the alternative normalization methods evaluated on Procgen, indicating that BatchNorm remains a highly effective normalization strategy. Whenever BatchNorm becomes unstable due to the train--evaluation mismatch, MDR mitigates this instability and achieves the strongest performance.

Overall, the results present a consistent picture across all three benchmarks. MDR provides its largest gains precisely when BatchNorm becomes unstable, while remaining largely performance-neutral otherwise. The consistency of these results across three benchmarks and two  architectures suggests that the benefits of MDR generalize across diverse environments and network backbones. 

\begin{figure*}[t!]
  \centering
\includegraphics[width=0.99\linewidth,keepaspectratio]{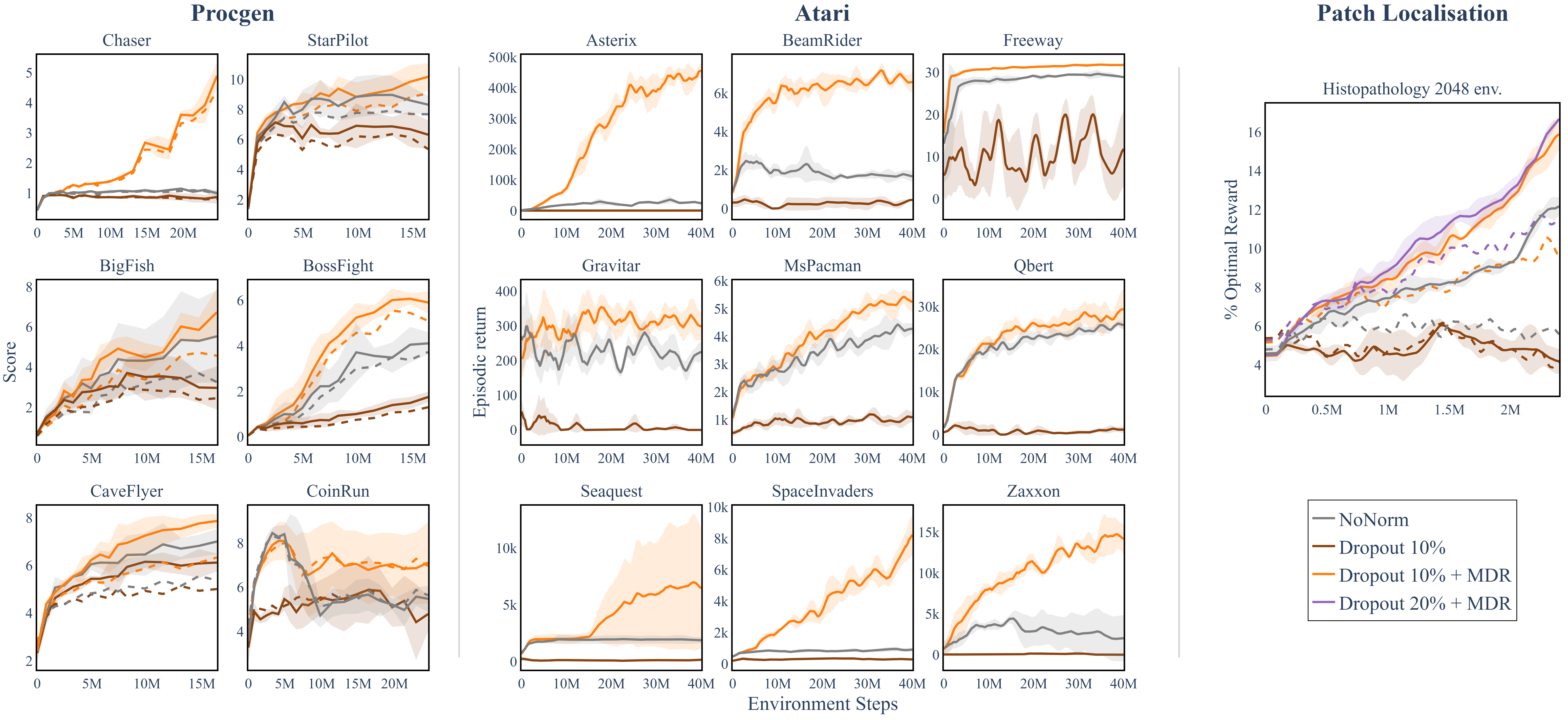}

    \caption{Dropout comparison across Procgen (left), Atari (center), and
    patch-localization (right) benchmarks. Solid lines denote training performance, dashed lines
    held-out evaluation. Dropout is applied at 10\%, with an additional 20\% variant
    for patch localization. Shaded regions denote one standard deviation across
three seeds.}
    \label{fig:dropout}
\end{figure*}

\subsection{Dropout with MDR}
We now evaluate whether rectification extends beyond BatchNorm to dropout. We report both training and held-out performance, with evaluation on 500 unseen Procgen levels and on unseen histopathology environments (Figure~\ref{fig:dropout}).

This experiment evaluates: (1) how a policy trained with dropout compares to the baseline; (2) whether MDR improves the stability and performance of dropout-based policies; and (3) whether the rectification procedure preserves the generalization benefits typically associated with dropout.

Figure~\ref{fig:dropout} shows that agents trained with dropout alone (Dropout) generally underperform the baseline (NoNorm) and exhibit reduced stability, with pronounced fluctuations and performance degradation.

In contrast, incorporating MDR substantially improves performance when using dropout across all three benchmarks. The effect is most pronounced on Atari, where dropout alone stagnates near baseline while Dropout+MDR improves steadily, most visibly on Asterix, BeamRider, Seaquest, SpaceInvaders, and Zaxxon. Across all benchmarks, Dropout+MDR outperforms the baseline, achieving higher and more stable returns; on Procgen and patch localization, where held-out evaluation is available, these gains also appear in test performance, suggesting that the improvements extend beyond the training environments.

One possible explanation of this improvement is the regularization effect of dropout, which encourages generalization even across training levels. A related phenomenon has been reported in prior work  \cite{CoinRun,Proce}, where increasing the number of training levels past a certain threshold leads to improved training performance. While the underlying mechanism differs, dropout in our case versus the number of environments in theirs, the observed behavior is consistent.

While Procgen demonstrates that Dropout+MDR improves both training and test performance, the resulting generalization gap is relatively small, making it difficult to isolate the effect of rectification on generalization alone. We therefore consider the histopathology patch-localization benchmark, which exhibits substantially stronger overfitting. As shown in Figure~\ref{fig:dropout} (right), Dropout+MDR improves both training and test performance relative to the baseline while reducing the generalization gap. These results provide evidence that the generalization benefits of dropout are retained under MDR in this setting, although a broader evaluation across additional overfitting-prone benchmarks would be needed to draw more general conclusions.

\subsection{Discussion}

Overall, the experiments show that MDR stabilizes mode-dependent layers across diverse domains. This is evident in both the BatchNorm and dropout settings, where MDR consistently improves or preserves performance without increasing the training budget because rectification simply reallocates the existing epoch budget

We evaluate two representative allocations corresponding to light and extensive rectification, although intermediate or more extreme splits (e.g., $(n{-}0.5,\,0.5)$) are possible. While this introduces an additional hyperparameter, these two settings proved sufficient across our experiments.

Our main limitation stems from the following observation: the two allocations do not have a consistent winner. In the dropout setting, the extensive variant tends to come out on top, whereas for BatchNorm the light variant does, with a few tasks in each setting where the opposite holds. This likely reflects the fact that different layers require different amounts of rectification, and that these requirements change as training progresses.

An adaptive allocation that adjusts rectification according to each layer's current needs is therefore a natural direction for future work, potentially guided by $\Delta_\epsilon$ or $\delta_r$. Extending the evaluation to continuous-control benchmarks and other trust-region algorithms is another promising direction.

\section{Conclusion}
\label{sec:Conclusion}
We showed that PPO training can be destabilized by mode-dependent layers such as BatchNorm and dropout. We traced this instability to a perturbation of PPO's trust region and introduced Mode-Dependent Rectification (MDR), a simple two-phase procedure that restores PPO's original trust-region constraint by periodically re-optimizing under deterministic layer behavior, without modifying the network architecture.

Across three benchmarks (Procgen, Atari, and a real-world patch-localization task) and two backbone architectures, MDR delivers its largest gains precisely where BatchNorm becomes unstable, while remaining competitive elsewhere. A corrected BatchNorm remains on par with normalization schemes that avoid the train--evaluation mismatch, and the same rectification extends to dropout. These results indicate that mode-dependent layers need not be discarded in PPO: their instability can instead be corrected while preserving their practical advantages.

\section{Acknowledgement}
This work has been supported by the French government, through the France 2030 investment plan managed by the Agence Nationale de la Recherche, as part of the ``UCA DS4H'' project, reference ANR-17-EURE-0004.

\bibliography{aaai2027}
\clearpage
\appendix

\input{app}

\end{document}

%% file: app.tex
\section{Benchmark}

\begin{figure*}
  \centering
    \includegraphics[width=0.95\linewidth]{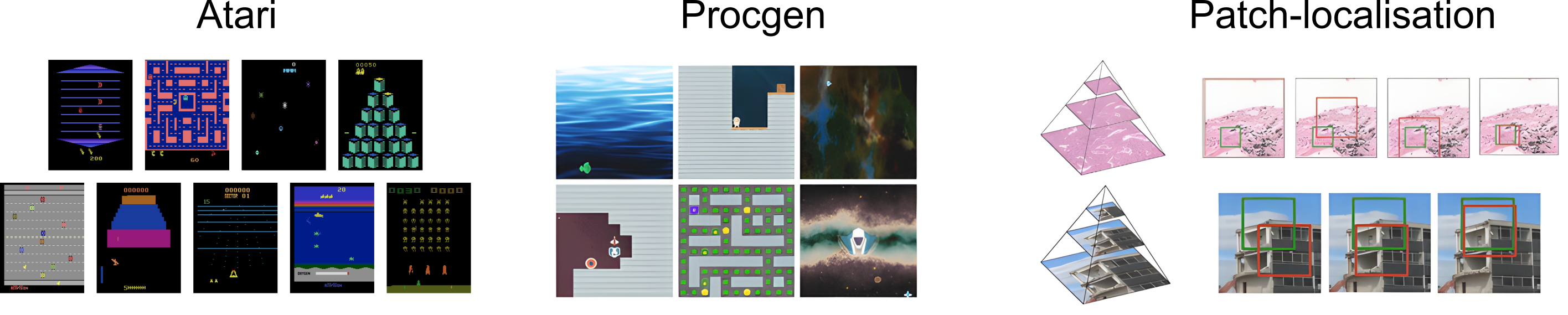}
    \caption{The three benchmarks used in our experiments. \textbf{Left:} Atari
    (ALE), showing the nine games used. \textbf{Center:} Procgen, six
    procedurally generated games with diverse visual styles. \textbf{Right:}
    patch localization, where the agent locates a target crop within a
    high-resolution image (histopathology, top; natural images, bottom); green
    boxes mark the target and red boxes the agent's current view.}
    \label{fig:training_pipeline}
\end{figure*}

This section provides a brief overview of the environments used in our experiments, illustrated in Figure~\ref{fig:training_pipeline}. We refer the reader to the original works for full environment specifications~\cite{CoinRun,Proce,bioimaging25,ale,ale2}.

\paragraph{Procgen.}
Procgen is a benchmark of procedurally generated video games designed to evaluate generalization in RL~\cite{CoinRun,Proce}. Each game supports "easy" and "hard" difficulty modes and defines environment-specific minimum and maximum achievable rewards. We evaluate six Procgen games:

\begin{itemize}
    \item \textbf{CoinRun}: the agent navigates to collect a coin while avoiding obstacles and enemies ($R_{\min}=5$, $R_{\max}=10$).
    \item \textbf{StarPilot}: a side-scrolling shooter game ($R_{\min}=2.5$, $R_{\max}=64$).
    \item \textbf{CaveFlyer}: the agent navigates a cave network to reach a friendly ship ($R_{\min}=3.5$, $R_{\max}=12$).
    \item \textbf{Chaser}: a pursuit-based game inspired by \textit{Ms. Pac-Man} ($R_{\min}=0.5$, $R_{\max}=13$).
    \item \textbf{BigFish}: the agent grows by consuming smaller fish while avoiding larger ones ($R_{\min}=1$, $R_{\max}=40$).
    \item \textbf{BossFight}: the agent must defeat a large enemy starship ($R_{\min}=0.5$, $R_{\max}=13$).
\end{itemize}

All Procgen environments share a $64 \times 64 \times 3$ RGB observation space and a discrete action space of 15 actions.

\paragraph{Atari.}
Atari, commonly referred to as the Arcade Learning Environment (ALE), is a standard video-game benchmark. We use the CleanRL implementation \cite{cleanrl} (\texttt{ppo\_atari\_envpool.py}) with $4\times84\times84$ observations; for further details, we refer the reader to their repository. The nine games used are described below:
\begin{itemize}
    \item \textbf{Asterix}: the agent moves horizontally and vertically to
    collect useful objects while avoiding lyres, which cost a life.
    \item \textbf{BeamRider}: the agent steers a forward-moving ship sideways
    between discrete positions to destroy enemy ships and dodge debris.
    \item \textbf{Freeway}: the agent guides a chicken across several lanes of
    traffic, scoring on each crossing.
    \item \textbf{Gravitar}: the agent pilots a small spacecraft through a solar system of planets, entering a side-view landscape upon reaching each one.
    \item \textbf{MsPacman}: the agent navigates a maze collecting pellets while
    avoiding ghosts.
    \item \textbf{Qbert}: the agent hops across a pyramid to change
    cube colors while avoiding enemies.
    \item \textbf{Seaquest}: the agent controls a submarine, rescuing divers and
    shooting enemies while managing a depleting oxygen supply.
    \item \textbf{SpaceInvaders}: the agent moves horizontally and fires at
    descending rows of aliens while dodging return fire.
    \item \textbf{Zaxxon}: the agent pilots a fighter and shoots enemies to stop the robot Zaxxon and its armies.
\end{itemize}
More information about the game could also be found here\footnote{https://ale.farama.org/environments/}. These nine games span a range of control schemes, objectives, and reward structures (from fast-paced shooting and evasion to maze navigation and resource management), exposing MDR to varied visual statistics and difficulty. Evaluating the full ALE suite is computationally prohibitive at our per-game budget; thus, we use a representative subset.

\paragraph{Patch Localization.}
Patch localization is a goal-conditioned visual navigation task \cite{bioimaging25} in which the agent must locate a target patch within a high-resolution image. Given an image of size $(m \times m)$, a target crop of size $(m/2 \times m/2)$ or $(m/4 \times m/4)$ is extracted and used as the goal. Both the full image and the target patch are resized to $112 \times 112$.

At each timestep, the agent observes a three-view input consisting of: (i) the target patch, (ii) a low-resolution view of the full image, and (iii) a local view centered at the agent’s current position. The resulting observation has shape $3 \times 112 \times 112 \times 3$. The agent acts in a discrete action space of seven actions, corresponding to navigation and zoom operations that allow it to explore the image and localize the target.

\section{Architecture Details and Initialization}
\subsection{Large IMPALA-CNN}

\begin{table}[t]
\centering
\caption{IMPALA-CNN backbone used for the Atari experiments, in its standard
and enlarged configurations.}
\label{tab:impala_architecture}
\renewcommand{\arraystretch}{1.2}
\begin{tabular}{lcc}
\toprule
\textbf{Stage} & \textbf{Standard} & \textbf{Enlarged (ours)} \\
\midrule
Input            & $(4,84,84)$ & $(4,84,84)$ \\
Conv sequence 1  & width $16$  & width $32$ \\
Conv sequence 2  & width $32$  & width $64$ \\
Conv sequence 3  & width $32$  & width $64$ \\
Conv sequence 4  & ---         & width $64$ \\
Embedding (FC)   & $256$       & $256$ \\
\midrule
Actor head       & \multicolumn{2}{c}{FC $\rightarrow$ Actions $(n)$} \\
Value head       & \multicolumn{2}{c}{FC $\rightarrow$ Value $(1)$} \\
\bottomrule
\end{tabular}
\end{table}

For the Atari experiments, we use the IMPALA-CNN~\cite{impala}, a convolutional residual architecture widely used in the visual RL literature~\cite{Proce}. We adopt the CleanRL implementation~\cite{cleanrl} and increase its depth and width relative to the standard configuration, adding a fourth convolutional sequence and doubling the channel widths. Unlike the ResNet backbone, the IMPALA-CNN is trained from random initialization. The enlarged architecture is shown in Table~\ref{tab:impala_architecture}.

\subsection{Shallow-ResNet18}

\begin{table}[t]
\centering
\caption{Shallow ResNet-18 architecture used in Procgen and Patch-localisation experiments (Block~4 removed).}
\label{tab:architecture}
\renewcommand{\arraystretch}{1.2}

\begin{tabular}{|c|c|}
\hline
\multicolumn{2}{|c|}{\textbf{Shared Backbone}} \\
\hline
\multicolumn{2}{|c|}{Input $(3,112,112,3)$} \\

\multicolumn{2}{|c|}{Conv1 $(3,56,56,64)$} \\

\multicolumn{2}{|c|}{Block 1 $(3,28,28,64)$} \\

\multicolumn{2}{|c|}{Block 2 $(3,14,14,128)$}\\

\multicolumn{2}{|c|}{Block 3 $(3,7,7,256)$}\\

\multicolumn{2}{|c|}{Block 4 (removed)} \\

\multicolumn{2}{|c|}{Average pooling $(3,1,1,256)$} \\
\hline
\textbf{Actor head} & \textbf{Value head} \\
\hline
Concatenate $(768)$ & Concatenate $(768)$  \\

FC  $(768)$ & FC  $(768)$ \\
FC  $(768)$ & FC  $(768)$ \\
\hline
Actions  $(7)$ & Value $(1)$ \\
\hline
\end{tabular}
\end{table}

One of the  architectures used throughout this work is derived from ResNet-18~\cite{resnet}, with the final residual block (Block~4) removed to reduce computational cost while preserving sufficient representational capacity. We refer to this variant as \emph{shallow ResNet-18}.

We primarily adopt ImageNet-pretrained weights. Beyond improving performance across all methods, this initialization is essential for making BatchNorm in evaluation mode a meaningful baseline.

\begin{figure}
  \centering
\includegraphics[width=0.9\linewidth,keepaspectratio]{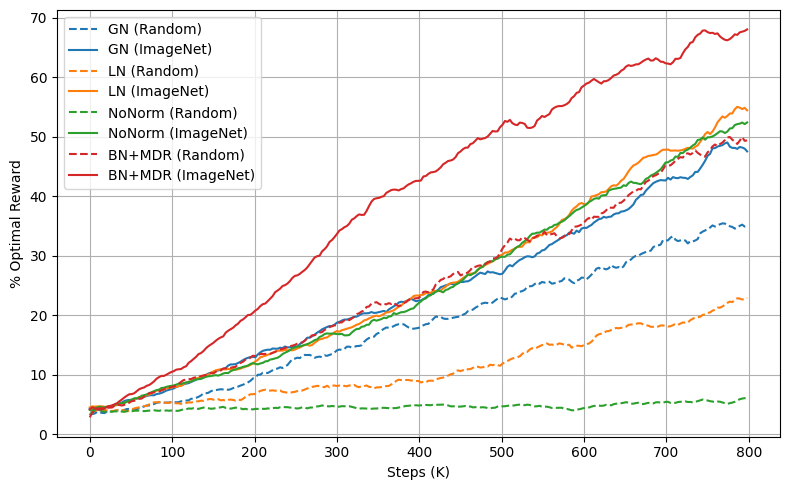}
    \caption{Effect of weight initialization on the histopathology
patch-localization task. Solid lines denote ImageNet-pretrained initialization,
dashed lines random initialization, for BN+MDR, LN, GN, and NoNorm. ImageNet
initialization substantially improves all methods; under both initializations,
BN+MDR achieves the highest performance.}
    \label{fig:architecture_motivation}
\end{figure}

Figure~\ref{fig:architecture_motivation} illustrates the effect of ImageNet initialization on BN+MDR, LN, GN, and NoNorm in the histopathology patch-localization task, showing a substantial improvement over random initialization. For completeness, the figure also compares normalization variants under random initialization, where BN+MDR achieves the highest performance despite an overall reduction across all methods.

Table~\ref{tab:architecture} summarizes the network architecture used for the actor and critic. Both networks share the same shallow ResNet backbone, followed by a two-layer MLP. For patch-localization tasks, three input views are processed independently using shared backbone weights and concatenated before the policy and value heads. For Procgen environments, the same backbone is used with a single input image, and the MLP dimensionality is adjusted accordingly.

\section{Hyperparameters}

We use the Adam optimizer and Generalized Advantage Estimation (GAE) throughout all experiments~\cite{adam,GAE}. A summary of hyperparameters is provided in Table~\ref{tab:hyperparameters}. Compared to the standard Procgen PPO configuration~\cite{Proce}, we disable reward normalization. For Atari, our
configuration differs from the CleanRL implementation in a few parameters chosen
for throughput: We raise the number of parallel environments from 8 to 32 while keeping 128 steps per environment, yielding a larger rollout of $4096$ transitions instead of $1024$. To keep the  batch size the same, we correspondingly increase the number of mini-batches from 4 to 16. These changes substantially improved throughput and reduced overall experiment time.

In the patch-localization task, advantage estimates and value targets are recomputed every three epochs. For a total of nine epochs per rollout, this results in three full MDR rounds per rollout.

\begin{table*}[t]
\centering
\caption{Hyperparameters used for Procgen and patch-localization experiments. For MDR, epochs are split across standard and rectification phases as described in text.}
\label{tab:hyperparameters}
\renewcommand{\arraystretch}{1.2}
\begin{tabular}{lccc}
\toprule
\textbf{Hyperparameter} & \textbf{Atari} & \textbf{Procgen} & \textbf{Patch localization} \\
\midrule
Rollout size $|D_k|$            & 4096  & 16384  & 3000  \\
Episode length                    & --- & 1000   & 20    \\
Parallel environments             & 32 & 64     & 4     \\
Epochs per update                 & 4 & 3      & 9     \\
Discount factor $\gamma$          & 0.99 & 0.999  & 0.99  \\
GAE parameter $\lambda$           & 0.95 & 0.95   & 0.95  \\
Value loss coefficient $c_1$     &  0.5 & 1.0    & 1.0   \\
Entropy coefficient $c_2$         &  $1\!\times\!10^{-2}$ & $1\!\times\!10^{-2}$ & $1\!\times\!10^{-4}$ \\
Batch size                      &  256  & 2048   & 128   \\
Learning rate                    & $2.5\!\times\!10^{-4}$  & $5\!\times\!10^{-4}$ & $4\!\times\!10^{-5}$ \\
Weight decay                      &  --- & $1\!\times\!10^{-4}$ & $1\!\times\!10^{-4}$ \\
Gradient clipping       & 0.5 & 1.0    & 1.0   \\
Advantage normalization          &   Yes & Yes    & Yes   \\
Input normalization (ImageNet mean/std) & No & No     & No    \\
Reward normalization                & No & No     & No    \\
\bottomrule
\end{tabular}
\end{table*}

\section{Reward Collapse}
\begin{figure*}
    \centering
    \includegraphics[width=0.94\linewidth]{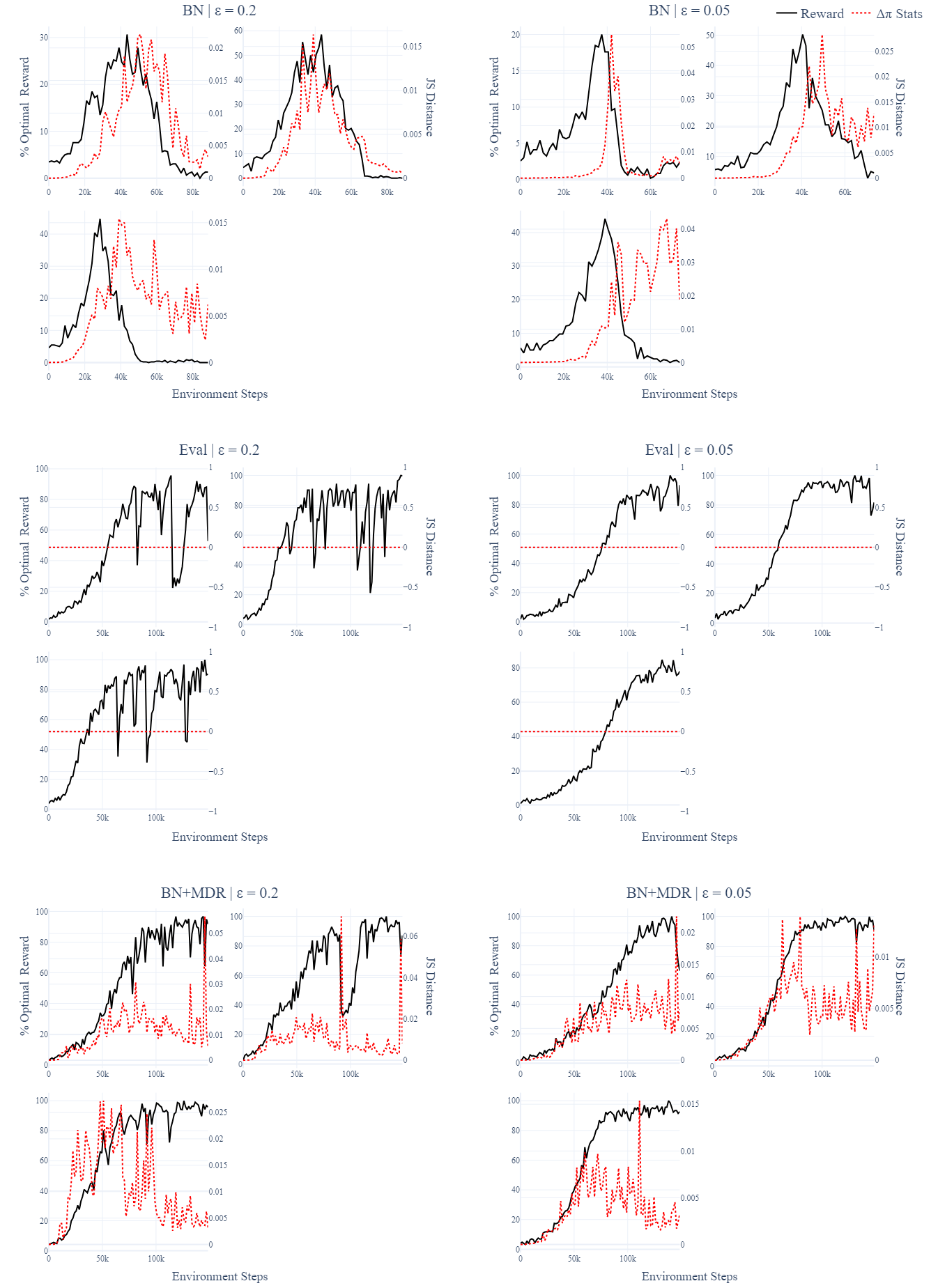}
    \caption{Reward and $\Delta \pi^-$ curves for agents trained in BN, Eval, and BN+MDR modes under two $\epsilon$ (clipping) values, over three seeds.}
    \label{fig:appendix-collapse}
\end{figure*}

Figure~\ref{fig:appendix-collapse} shows reward alongside the policy-mismatch
measure $\Delta\pi^-$ for agents trained under different BatchNorm
configurations (BN, Eval, and BN+MDR) using shallow Resnet-18 at two clipping values
($\epsilon \in \{0.2, 0.05\}$), over three seeds. To make the collapse
dynamics easier to observe, these runs use fewer transitions per dataset $D_k$,
fewer updates per step, and only four histopathology environments, which
amplifies distributional shift.

\noindent\textbf{BN.} Reward initially increases while $\Delta\pi^-$ grows, until
reward collapses abruptly. After collapse, $\Delta\pi^-$ remains elevated,
reflecting the persistent mismatch between pre- and post-collapse state
distributions. Collapse occurs at both clipping values.

\noindent\textbf{Eval.} With BatchNorm fixed in evaluation mode, training is
substantially more stable and $\Delta\pi^- = 0$ by definition.

\noindent\textbf{BN+MDR.} Here temporary reward drops are followed by a spike in $\Delta \pi^-$. Crucially, these spikes decay rapidly, and the agent recovers without entering a collapse regime. This behavior is especially pronounced for $\epsilon=0.2$, where BN fails but BN+MDR recovers. Lowering $\epsilon$ further reduces reward fluctuations.

\begin{figure*}[h]
    \centering
    \includegraphics[width=0.99\linewidth]{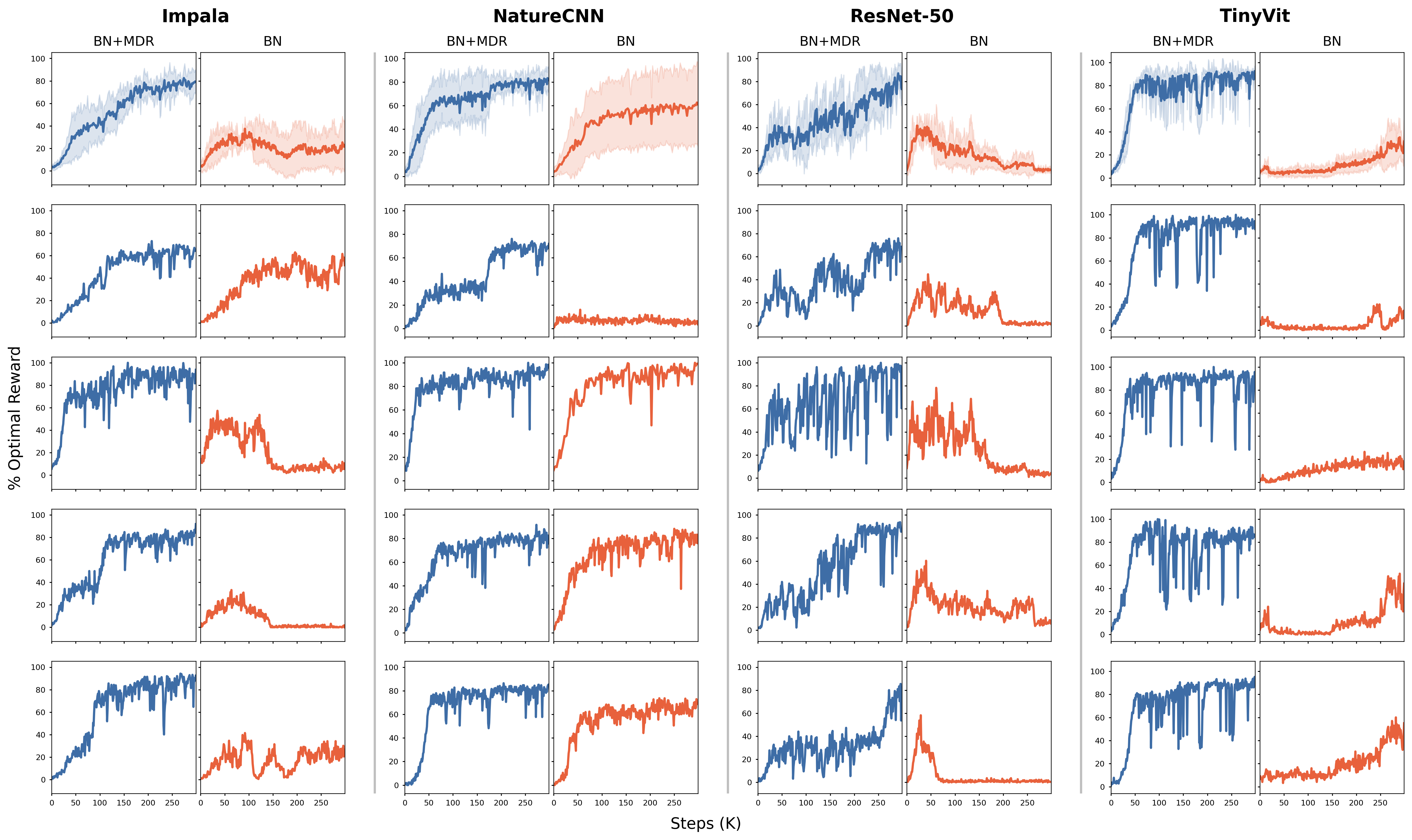}
    \caption{MDR across network architectures, on the histopathology patch-localization task. Each column pair compares BN+MDR (blue) against BN (red) for one architecture: Impala-CNN, NatureCNN, ResNet-50, and TinyViT. The top row shows the mean across four seeds (shaded: one standard deviation); each row below shows an individual seed. The benefit of MDR is consistent across all four architectures: BN training is unstable and often collapses, whereas BN+MDR remains stable and reaches high reward regardless of backbone.}
    \label{fig:arch_comparison}
\end{figure*}

\paragraph{MDR across architectures}
To further assess whether the benefits of MDR are specific to a particular
backbone, we repeat the histopathology experiment (discussed in the previous paragraph) across four architectures spanning a range of designs and capacities: NatureCNN \cite{DQN}, Impala-CNN, ResNet-50, and TinyViT (11M) \cite{tinyVit}, the last is a hybrid vision transformer employing both BatchNorm and DropPath (stochastic depth), giving it two distinct mode-dependent
components rather than one. For each architecture, we compare BN+MDR against standard BN over four seeds (Figure~\ref{fig:arch_comparison}).

Across architectures, BN+MDR is consistently stable and reaches high reward, whereas standard BN is not. The form of BN instability, however, varies with the backbone: Impala-CNN and ResNet-50 exhibit clear collapse on several seeds, while NatureCNN is comparatively more stable, which we believe stems from its shallower architecture stacking fewer mode-dependent layers. TinyViT behaves the least regularly of the four. This may relate to its second
mode-dependent layer or to its distinct transformer-based design. In all cases, BN+MDR avoids the instability seen under standard BN, indicating that rectification addresses this failure across diverse architectures and is not tied to a single backbone.

\section{Rectification and Entropy}

\begin{figure}
  \centering
\includegraphics[width=0.99\linewidth,keepaspectratio]{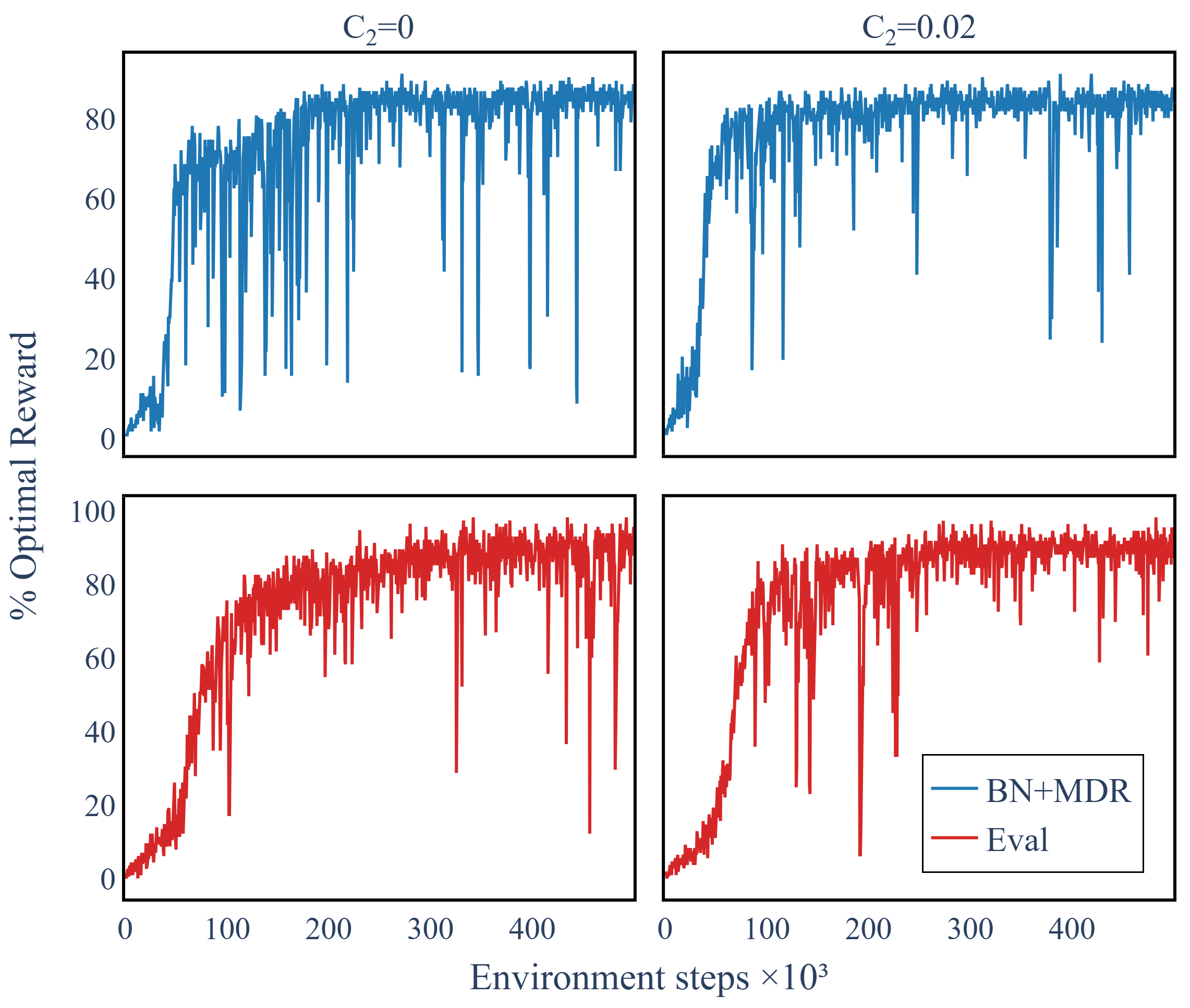}

\caption{Effect of entropy regularization under rectification. Top row: BN+MDR; bottom row: Eval. Columns correspond to different entropy coefficients. Removing entropy leads to increased reward fluctuations under BN+MDR, while having a limited effect under Eval.}

    \label{fig:fluctuation}
\end{figure}

The entropy bonus in the PPO objective provides a corrective mechanism during rectification: when stochastic perturbations in the standard phase drive the policy beyond the original trust-region boundary, yielding an overly deterministic solution, the entropy term favors higher-entropy alternatives in the rectification phase.

This can be formalized as follows. Consider a training phase in which the clipped PPO objective is evaluated with a perturbed bound $\epsilon + \Delta\epsilon$, producing an excessively deterministic policy $\pi$ that saturates the original clipping region. There may then exist a more stochastic policy $\pi'$ such that
\begin{equation}
\begin{gathered}
L^{\text{clip}}(\pi', \epsilon + \Delta\epsilon) < L^{\text{clip}}(\pi, \epsilon + \Delta\epsilon)\\
L^{\text{clip}}(\pi', \epsilon) = L^{\text{clip}}(\pi, \epsilon)\\
\text{and}\quad \mathcal{S}(\pi') > \mathcal{S}(\pi),
\end{gathered}
\label{eq:proof}
\end{equation}
where $\mathcal{S}(\pi)$ denotes policy entropy. Through the entropy term, the PPO objective in the rectification phase favors $\pi'$ over $\pi$, restoring stochasticity and stabilizing optimization.

We show this empirically on the histopathology patch-localization task. To accentuate instability, all agents use fewer transitions per dataset $D_k$, and we compare BN+MDR and Eval trained with and without the entropy bonus. Because disabling entropy can lead to suboptimal behavior, we do not report the mean across seeds; instead, for each configuration, we take the two seeds with the largest reward fluctuations and aggregate them using a minimum operator to highlight worst-case instability (Figure~\ref{fig:fluctuation}). Under Eval,
removing entropy has little effect on stability. Under BN+MDR, reward fluctuations increase substantially, suggesting that entropy regularization plays a stronger stabilizing role in correcting perturbations. 

\begin{figure*}
  \centering
\includegraphics[width=0.99\linewidth,keepaspectratio]{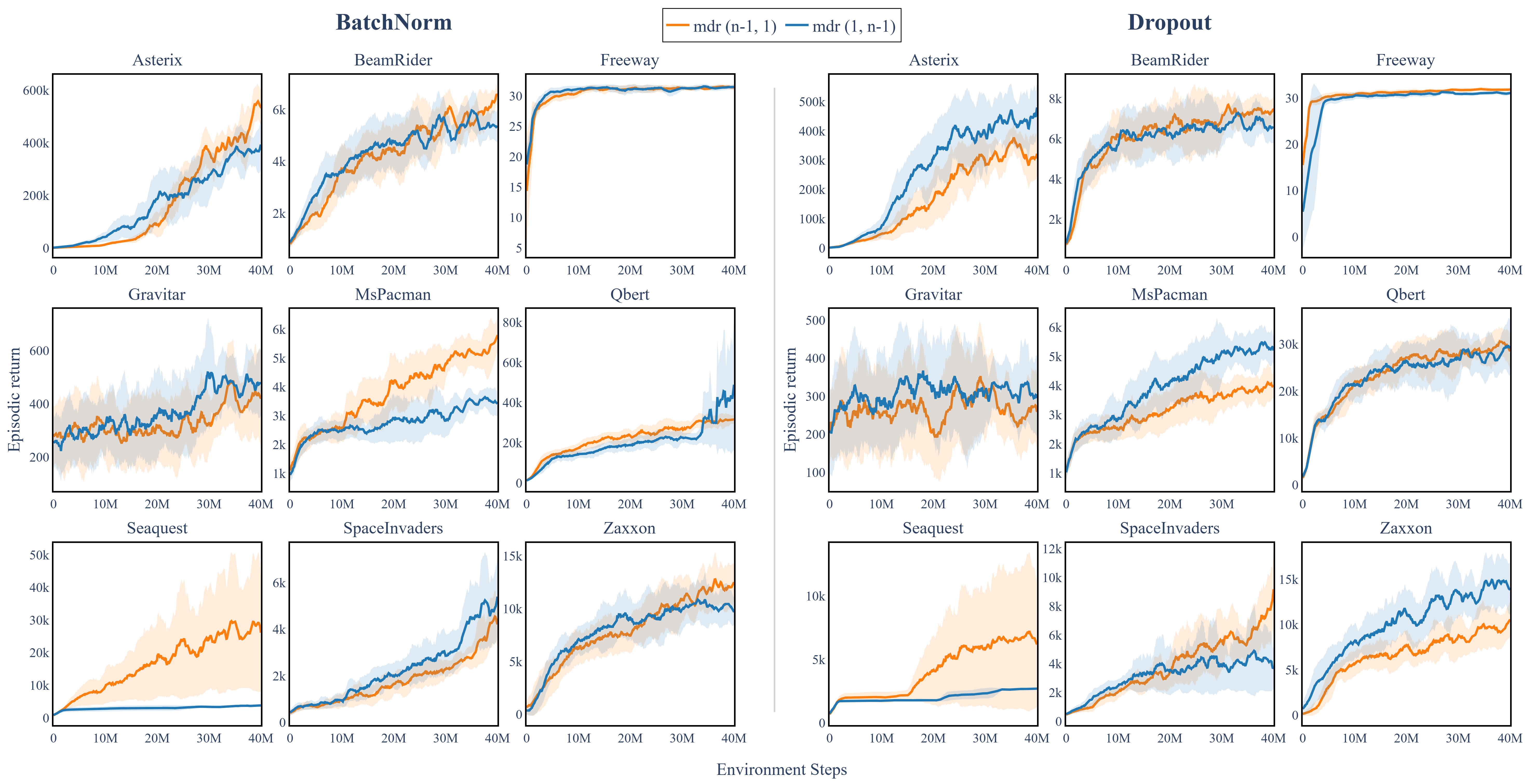}

\caption{Comparison of the two MDR allocations, light $(n{-}1,1)$ and extensive
$(1,n{-}1)$, on the nine Atari games, for BatchNorm (left) and dropout (right).
Shaded regions denote one standard deviation across three seeds.}

    \label{fig:Extra}
\end{figure*}